\documentclass[runningheads]{llncs}
\usepackage{graphicx}
\begin{document}
\title{MDR Cluster-Debias: A Nonlinear Word Embedding Debiasing Pipeline}
%
%
\author{Yuhao Du\orcidID{0000-0002-2474-8529} \and
Kenneth Joseph\orcidID{0000-0003-2233-3976}}

%
%
\institute{University at Buffalo, Buffalo NY 14260, USA  \\ \email{\{yuhaodu,kjoseph\}@buffalo.edu}}
\maketitle              
\begin{abstract}
Existing methods for debiasing word embeddings often do so only superficially, in that words that are stereotypically associated with, e.g., a particular gender in the original embedding space can still be clustered together in the debiased space. However, there has yet to be a study that explores why this residual clustering exists, and how it might be addressed. The present work fills this gap.  We identify two potential reasons for which residual bias exists and develop a new pipeline, MDR Cluster-Debias, to mitigate this bias. We explore the strengths and weaknesses of our method, finding that it significantly outperforms other existing debiasing approaches on a variety of upstream bias tests but achieves limited improvement on decreasing gender bias in a downstream task. This indicates that word embeddings encode gender bias in still other ways, not necessarily captured by upstream tests.

\end{abstract}
\section{Introduction}

A literature has rapidly developed around the question of how to identify, characterize, and remove bias from (``debias'') word embeddings. Attempts to do so are critical in ensuring that real-world applications of natural language processing (NLP) do not cause unexpected harm.  For example, word embeddings that reflect stereotypical and/or prejudicial social norms might be used as input to other algorithms that, e.g., rank men higher than more qualified women for job searches of particular occupations \cite{Caliskan_2017}.

However, recent work has raised questions about existing efforts to measure biases in word embeddings, and our ability to debias them. With respect to measurement, Ethayarajh et al. \cite{ethayarajh-etal-2019-understanding} provide both empirical and theoretical evidence that the most common method of measuring bias in word embeddings, the \emph{Word Embedding Association Test} (WEAT), provides unreliable measures of practical and statistical significance. With respect to debiasing, Gonen and Goldberg \cite{Gonen2019LipstickOA} provide experimental evidence that male-stereotyped words are still easily distinguishable from female-stereotyped words after running two of the most well-known methods for debiasing, the Hard-Debias method of \cite{Bolukbasi:2016:MCP:3157382.3157584} and the Gender Neutral GloVe (GN-Glove) approach \cite{Zhao2018LearningGW}.  

These two debiasing methods, like nearly all others, operate under the assumption that social biases in word embeddings can be defined as a specific direction (or in some cases, a subspace) of the embedding space. This direction is characterized by the difference between sets of bias-defining words. For example, the Hard-Debiasing method of Bolukbasi et al.  \cite{Bolukbasi:2016:MCP:3157382.3157584} approach works roughly as follows for gender debiasing. First, a ``gender direction'' is identified by using differences in the embedding space between sets of gender-paired words, e.g. ``man'' and ``woman''.  This direction is then essentially removed from all other words,\footnote{Except for ``gender definitional'' words like ``king and queen''} with the idea being that gender will no longer be represented by the embeddings of the remaining terms because all gender information, contained on the gender direction, is now gone.


What Gonen and Goldberg show is that while this approach removes some forms of gender information, one can still easily pick up gender stereotypes in word embedding space based on different, but equally valid, definitions of bias metrics. Inspired by their work, we propose a new debiasing procedure which combines a post-processing step, introduced in \cite{Hasan2017WordRV}, to unfold manifolds in high word embedding space, followed by a simple linear debiasing approach, Cluster-Debias, that finds a better direction along which to remove bias to address these cluster-based bias measures.

We evaluate our debiasing approach for several ``upstream'' tasks, including bias tests and word similarity tests. In addition, we ask, what does this means for downstream performance on a standard NLP task?  We compare embeddings debiased using our approach with the approaches of \cite{Zhao2018LearningGW} and \cite{Bolukbasi:2016:MCP:3157382.3157584} on a coreference resolution task and a sentiment analysis task.  Through these efforts, the present work makes the following contributions to the literature:\footnote{Code and data to replicate our work are at https://github.com/yuhaodu/MDR-Cluster-Debias.git.}
\begin{itemize}
    \item We find evidence that debiased embedding clusters are partially due to manifold structure in high dimensional word embedding space. 
    \item We introduce a new pipeline to perform debiasing, and show it can reduce the clustering-based word embedding bias measures introduced by \cite{Gonen2019LipstickOA}.
    \item However, despite significant upstream improvements, our approach does not significantly decrease bias in the downstream task of coreference resolution.
\end{itemize}


\section{Related Work}

\subsection{Bias Definition}\label{sec:def}

Critical to debiasing is how bias is actually defined. The vast majority of works use a directional definition.  Under this approach, a single direction in the embedding space defines a particular bias that the authors expect to exist, e.g. the ``gender direction''.
Detractors of the directional definition of bias, like Gonen and Goldberg, have argued that it is inappropriate, because a single gender (or race, etc.) dimension may not capture all forms of bias encoded in the data. This issue has led others to define bias in terms of clustering, or word proximity. The idea is that the removal of a single dimension, or subspace \cite{Manzini2019BlackIT}, is not sufficient to remove bias, in that one can easily identify terms that are close to the opposing seed terms in the original embeddings in the unbiased embeddings as well. Because of this, protected information (e.g. gender) can potentially leak to machine learning algorithms in downstream tasks. The primary contribution of our work is to propose a debiasing pipeline to resolve these \emph{cluster-based biases}.

\subsection{Debiasing Word Embeddings}

%


Several methods have been proposed to remove social biases from various kinds of NLP methods; see Sun et al. \cite{sun_mitigating_2019} for a recent review on gender specifically. Bolukbasi et al. \cite{Bolukbasi:2016:MCP:3157382.3157584} proposed two methods for word embeddings specifically, Hard Debiasing and Soft Debiasing.  These methods remove gender neutral words' projection over a gender space defined by gender definitional words. Zhao et al. \cite{Zhao2018LearningGW} modify the GloVe algorithm to train debiased embeddings directly from a co-occurrence matrix by adding constraints in the training objective of GloVe \cite{pennington2014glove} to force gender neutral words perpendicular to some gender space.  


Except these two seminal works, several others have proposed novel methods for debiasing. Most of these have been extensions of the hard-debiasing method. \cite{Manzini2019BlackIT} extend the hard debias method to the multi-class setting,  \cite{ethayarajh-etal-2019-understanding} improve the way gender-biased words are selected, and \cite{dev2019attenuating} propose simpler versions of the algorithm and the use of names as a means of identifying directions in the space that represent social biases.  One exception is the work of Kaneko et al. \cite{kaneko_gender-preserving_2019}, who propose an autoencoder based method which is able to project current word embedding into another space which preserves the word semantic information while removing the gender bias. However, their work still evaluates results using a directional approach. The present work extends current debiasing algorithms in terms of debiasing based on recently identified cluster-based bias definition.

\section{Our Debiasing Pipeline}


We base our debiasing pipeline on pretrained GloVe embeddings \cite{pennington2014glove}; however, the approach generalizes to any other pretrained embedding. The pipeline contains two parts: the first is a post-processing procedure, which is used to re-embed original word vectors into a new space via a manifold learning algorithm. The second is the application of a direction-based debiasing method to remove gender information in the re-embedded word vectors. 



\subsection{Post-Processing Procedure}\label{sec:MDR} 

As a post-processing procedure, we use Manifold Dimensionality Retention (MDR) from  \cite{Hasan2017WordRV}.  Hasan et al. \cite{Hasan2017WordRV} are motivated by the observation that word embeddings slightly underestimate the similarity between similar words and overestimate the similarity between distant words. This indicates that word embedding space contains non-linear manifold structure. Thus, they propose the MDR to unfold the manifold structure to improve word representation and results show that re-embedded word embeddings achieve better performance in word similarity tests. Inspired by their observation and Gonen and Goldberg's observation that gendered words are easily separated by a non-linear SVM method, we believe non-linear manifold structure in the word embedding space could potentially prevent linear directional based debias method from mitigating gender bias. Thus, we apply MDR as a post-processing procedure. 

In MDR, we start from an original embedding space with vectors ordered by words frequencies. We then carry out the following steps: 
\begin{enumerate}
    \item Select a sample window of vectors that are used to learn the manifold. 
    \item Fit a manifold learning model to the selected sample using Locally Linear Embedding (LLE)  \cite{Roweis2000NonlinearDR}.
    \item The resulting fitted model is then used to transform  all the word vectors in the original space to the new re-embedding space.
\end{enumerate}

In Step 1, a sample window is sliding on the word vectors ordered by word frequencies. The window length $L$ and window start $S$ of the sample window are hyper-parameters. Additional, $S$ will decide the computational complexity on manifold learning. As shown in prior work \cite{Gong2018FRAGEFW}, trained word embeddings are biased toward word frequency. In order to keep learned manifold from skewing towards high frequency or low frequency words, we select $S$ as $5000$. For the choice of $L$, we choose $1000$ following the suggestion introduced in the prior work \cite{Hasan2017WordRV}. Selections of these two parameters work well in terms of preserving semantic information in word embeddings which is shown later in Section~\ref{sec:results}.  

\subsection{Cluster-Debias}\label{sec:clust}
Gonen and Goldberg show that, after \emph{debiasing}, one can still easily cluster biased words using linear K-means clustering method. We hypothesize that this observation is due to a mismatch between the direction that previous debiasing method removes and that gender bias lies along. Thus, we propose a simple approach that incorporates a cluster-based definition of bias to perform debiasing. 
The procedure carried out by Cluster-Debias is as follows:
\begin{enumerate}
    \item Identify, via a particular word pair $a$ and $b$ (e.g. ``he'' and ``she''), the form of bias to be addressed. 
    \item Identify the bias subspace by $D_{bias} = E_{a} - E_{b}$, where $E_w$ represents the word vector of word $w$. 
    \item Calculate the bias of word vectors along $D_{bias}$ following the methods in \cite{Bolukbasi:2016:MCP:3157382.3157584}
    \item Select the top $k$ most biased words, i.e. the $k$ nearest neighbors to $a$ and $b$
    \item Apply PCA over these $2k$ word vectors and extract the first principle component $D_{pc}$. 
    \item Debias all word vectors $E_w$s by removing $D_{pc}$ from them. This can be expressed as $E^{'}_{w} = E_{w} - \langle E_{w}\,,D_{pc}\rangle \cdot D_{pc} $.
\end{enumerate}

At a high level, our approach retains much of the logic from Bolukbasi et al.  \cite{Bolukbasi:2016:MCP:3157382.3157584}. However, instead of assuming that the gender direction is aligned with the word pairing(s) we identify, we instead assume that this direction can be better identified by incorporating information from the distribution of the word vectors that are proximal to $a$ and $b$.  We therefore assume, based on the observations of Gonen and Goldberg \cite{Gonen2019LipstickOA}, that it is more appropriate to select a direction based on the clustered structure of the embedding space around the words of interest, rather than on those words themselves. Note that it is not guaranteed that the Cluster-Debias approach will overcome the issues with Hard-Debias.  Like the Hard-Debias method, we remove only a single dimension from the embedding space. This direction is simply more informed by clustering structure than the prior work.
Here, we focus on comparing to prior work, and so consider gender. Thus $a$ and $b$ are ``he'' and ``she'', respectively. Additionally, we set $k= 1000$ for all experiments below.

\section{Evaluation Methods}

There is, of course, a tradeoff between removing gender information and maintaining other forms of semantic information that are useful for downstream tasks. As such, we evaluate embeddings from bias-based evaluation measures and semantic-based evaluation measures. In addition, we are also interested in whether or not upstream evaluation results can be transferred to downstream tasks. As such, our evaluation is carried out along two dimensions --  bias-based versus semantic-based and upstream versus downstream. 




As an upstream measure of semantics, we focus on semantic similarity-based measures. We compute the cosine similarity between word embeddings and measure Spearman correlation between human similarity rating and cosine similarity for the same semantic relatedness datasets used to evaluate biased embeddings in prior work \cite{Zablocki2017LearningMW}.
For downstream evaluation, we identify two NLP tasks -- coreference resolution and sentiment analysis. To build coreference resolution models, we use the coreference resolution system proposed in \cite{Lee_2018}. We apply the original parameter settings for the model and train each model for 100K iterations and evaluate models with respect to their performance on the standard OntoNote v5 dataset \cite{OntoNotes}. For sentiment analysis, we train an LSTM with 100 hidden units on the Stanford IMDB movie review dataset \cite{maas-EtAl:2011:ACL-HLT2011} and we also leverage the model \cite{kim-2014-convolutional} to train a binary classifier on the MR dataset of short movie reviews \cite{pang-lee-2005-seeing}.




For bias-based evaluation, we use the same six cluster-based bias measures that are proposed by Gonen and Goldberg as our upstream bias-based evaluation tasks. The first one we call \emph{Kmeans Accuracy}. We first select the top 500 nearest neighbors to the terms ``he'' and ``she'' in the original embedding space. We then check the accuracy of alignment between gendered words and clusters identified by Kmeans. The second one we call \emph{SVM Accuracy}. We consider the 5000 most biased words (2500 from each gender) in the original embedding space. \emph{After debiasing}, we check the accuracy of a RBF-kernel SVM trained on a random sample of 1000 of these words (500 from each gender) predicting gender bias of the remaining 4000. The third one we call \emph{Correlation Profession}. We extract the list of professions used in \cite{Bolukbasi:2016:MCP:3157382.3157584} and compare the correlation between the percentage of male/female socially biased words among the $k$ nearest neighbors of the professions and their directional bias in the original embedding space. For three metrics listed above, lower scores indicate better debiasing results. The rest three are gender-related association experiments called \emph{WEAT} introduced in \cite{Caliskan_2017}. Three experiments evaluate the associations between female/male and family and career words, arts and mathematics words, arts and science words respectively. For these tests, a higher p-value means lower association which indicates better debiasing results.  


For downstream evaluation of bias, We again leverage the coreference model that is trained following the procedure introduced above as our model for bias-based tests. The difference between here and there is that we compare performance using the gendered coreference resolution dataset WinoBias developed by Zhao et al. \cite{zhao-etal-2018-gender}. The testing portion of the WinoBias dataset evaluates the extent to which a coreference resolution model exhibits gender stereotyping by assessing the degree to which it applies gender stereotypical pronouns to individuals described using a set of gender-associated occupations. They create two different datasets—“anti-stereotype”, in which gender associations are reversed (e.g. “The secretary ... he”), and “pro-stereotype”, in which gender associations are retained (e.g. “The secretary ... she”). Differences in performance between the two datasets are used as an indicator of gender bias in the coreference dataset and gender bias in coreference algorithm.

\section{Experiments}



We train GloVe \cite{pennington2014glove} on a 2017 dump of English Wikipedia to obtain pre-trained 300-dimensional word embeddings for 362179 words. We then create several baselines and word embeddings debiased by our proposed methods:

\textbf{GloVe}: is the pretrained word embedding introduced above. This baseline denotes a non-debiased version of the word embeddings. 

\textbf{Hard-GloVe}: We apply hard-debiasing \cite{Bolukbasi:2016:MCP:3157382.3157584} method by using released code \footnote{\url{https://github.com/tolga-b/debiaswe}} to our pretrained GloVe word embedding and obtain a hard-debiased version of the pretrained GloVE embeddings.


\textbf{GN-GloVe}: We apply the code \footnote{\url{https://github.com/uclanlp/gn\textunderscore glove}} from original authors of GN-GloVe \cite{Zhao2018LearningGW} and train our own version of GN-GloVe. 


\textbf{Cluster-GloVe}: We apply Cluster-Debiased method to our pretrained GloVe embeddings to obtain debiased GloVe embeddings. 

\textbf{MDR-GloVe}: We apply our Post-Processing Procedure MDR on pretrained GloVe embeddings. 

\textbf{MDR-Cluster}: We apply the proposed Post-Processing Procedure on pretrained GloVe embedding and then use Cluster-debias method to debias it.


\textbf{MDR-Hard}: To test whether our Post-Processing Procedure works for other debias methods. We apply proposed Post-Processing Procedure on pretrained GloVe embedding and then use Hard-debias method to debias it.

\begin{table}[t]
\small
    \centering
    \caption{Results for our upstream bias evaluations.  The first three rows are extracted directly from prior work \cite{Gonen2019LipstickOA}. The last four rows are the debiased word embeddings using our proposed pipeline. Bolded results are the best-performing in each column according to the given metric. }
    \label{tab:h1}
    
\resizebox{.95\columnwidth}{!}{
\begin{tabular}{p{2.cm}p{1.3cm}p{1.1cm}p{1cm}p{2.2cm}p{2cm}p{2cm}}

{\bf Embedding} & {\bf Kmeans Acc.} & {\bf Corr. Prof.} & {\bf SVM Acc.} & {\bf Work/Family P-val} & {\bf  Math/Art 1 P-val} & {\bf Math/Art 2 P-val} \\ \hline
{\bf Original GloVe} & 0.999 & 0.820 & .99 &  & & \\ \hline
{\bf Hard-GloVe} & 0.925 & 0.606 & .89 & \textless .0001 & \textless .0001  & .0467 \\ \hline
{\bf GN-GloVe} & 0.856 & 0.792 & .97 & \textless .0001  & \textless .0001 & \textless .0001  \\ \hline
{\bf Cluster-GloVe} &{\bf 0.53} & 0.74 & 0.80 & \textless .0001  & \textbf{0.76} & 0.20 \\ \hline
{\bf MDR-GloVe} & 1.000 & 0.88 & 0.99 & \textless .0001  & 0.09 &  0.03 \\ \hline
{\bf MDR-Cluster} &0.556 & {\bf 0.38} & {\bf 0.518} & 0.00015  & 0.43 &  0.26 \\ \hline
{\bf MDR-Hard} &0.915 & {\bf 0.38} & 0.86 & {\bf 0.002}  & 0.42 & {\bf 0.51} \\ \hline
\end{tabular}}
\end{table}
\subsection{Results}\label{sec:results}


\subsubsection{Upstream Cluster-based Bias Test}\label{sec:eva_cluster}

Table~\ref{tab:h1} displays results from our upstream bias evaluations, and shows that our new debiasing strategies significantly improve over prior work. Results can be summarized as follows:
\begin{enumerate}
\item Cluster-GloVe outperforms GN-Glove and Hard-GloVe on all cluster-bias based tests because of following reasons. First, Post-debias Cluster Accuracy of Cluster-GloVe is 0.53, which means that debiased gendered words are not separable by K-means. Cluster-GloVe is also the most difficult to classify post-hoc using an SVM (it has lowest SVM Classifier Accuracy). And it shows no gender bias on two of the three WEAT tests (p-values of last two columns show no significant results). With respect to deficiencies in the Cluster-GloVe, embeddings still are highly separable by SVM, and as evidenced from the correlational professions experiment and the Work/Family WEAT, retain gender stereotypes for occupations.

\item Our Post-Processing method is able to help not only the Cluster-Debias method but also HardDebias. MDR-Cluster and MDR-Hard outperform Cluster-GloVe and Hard-GloVe respectively. 

\item MDR-Cluster achieves the best overall performance and is the only method that prevents SVM from classifying gender stereotyped words, which validates the efficiency of our debias pipeline. But MDR-Cluster still struggles with work/family associations WEAT test.


\end{enumerate}


These results provide two insights into the observations of Gonen and Goldberg. First, Cluster-GloVe, as a directional based debias method, out-performs Hard-GloVe (also a directional based debias method) in terms of removing post-bias clusters identified by K-Means. This observation suggests that there is a mismatch between the direction that gendered words distribute along and the direction that prior debias methods remove. Second,  the fact that Cluster-GloVe removes post-debias clusters identified by K-Means, but not non-linear SVM, and that MDR-Cluster removes both, suggests that manifold structure in the word embedding space is able to leak protected gender information to non-linear method (e.g. SVM). That validates our decision on using MDR to unfold manifold structure in the word embedding space as our post-processing step. 





\subsubsection{Upstream Semantic Similarity and Relatedness} \label{sec:sim}

We find that, compared with others, word embeddings debiased by our proposed pipeline achieve as-good or higher performance on most benchmark datasets for our upstream semantic test. The most critical comparison is to the original embeddings, where on average, MDR-Cluster achieves 60.2 Pearson correlation with the ground truth ratings on the five benchmark tasks, while GloVe achieves 56.2. This indicates that, according to the word similarity test metric, our proposed debiased pipeline can keep or amplify the semantic information in the original word embeddings.\footnote{Full result tables are available at https://github.com/yuhaodu/MDR-Cluster-Debias.git}


\subsubsection{Downstream Results - Coreference Resolution}
\begin{table*}[t]
    \small
    \centering
    \caption{Performance on co-reference resolution task for models trained using the given embedding. All performance scores are given as F1 scores. Bolded results are the best-performing in each column.}
    \label{tab:my_label}
    \resizebox{.95\columnwidth}{!}{
    \begin{tabular}{p{2.3cm}|p{2.cm}|p{2.cm}|p{2cm}|p{1.7cm}|p{1.5cm}}
    {\bf Model} & {\bf OntoNote} & {\bf Anti-Stereotype} & {\bf Pro-Stereotype} & {\bf WinoBias Mean} &{\bf WinoBias Diff.} \\ \hline
    GloVe & 72.49 & 60.995  & 81.535  & 71.265 & 20.54\\ 
    Hard-Debias  & 71.87 & 63.27 & 77.69 & 70.48 & 14.42\\ 
    GN-GloVe & \textbf{72.69} & 65.47 & 81.415  & 73.4425  & 15.945 \\
    Cluster-debias & 71.94 & 63.685 & 82.125 & 72.905 & 18.44\\ 
    MDR & 71.93 & 65.715 & \textbf{83.59} & \textbf{74.6525} & 17.875\\ 
    MDR-Hard & 71.70 & 66.18 & 79.78 & 72.98 & \textbf{13.6}\\ 
    MDR-Cluster & 72.01 & \textbf{66.73} & 80.66 & 72.69 & 13.93\\ 
    \end{tabular}}
\end{table*}

Table~\ref{tab:my_label} shows the performance difference between coreference resolution algorithms based on the different debiased GloVe embeddings. Among the embeddings considered, we find that MDR-Hard and MDR-Cluster show the best performance on the WinoBias datasets on WinoBias Difference and the Anti-Stereotype metric.  This suggests that in addition to showing improvements on the tasks studied by Gonen and Goldberg, MDR-Hard and MDR-Cluster methods we propose can attenuate more protected gender information in the word embeddings.  We further can find that different methods' performances are similar on OntoNote dataset which indicates that our debias pipeline doesn't deprecate the semantic information that are essential for coreference resolution. However, differences between the various debiasing strategies are limited, compared to their overall difference from the unbiased, original embeddings.

\subsubsection{Downstream Results - Sentiment Analysis}

Finally, we find that the models trained on MDR-cluster achieve a similar accuracy on sentiment classification on the MR dataset. However, using the MDR-Cluster embeddings, accuracy on IMDB dataset is 68.5\% (95 \% confidence interval [68.2\%, 68.9\%]), while using GloVe embeddings, accuracy is 80.9\% ([80.5\%,81.3\%]). This drop in sentiment analysis performance is indicative that debiasing along certain dimensions of stereotyping (e.g., gender), may have important downstream effects. Although the present work focused on addressing issues raised in \cite{Gonen2019LipstickOA}, this finding on an  important downstream task suggests future work is needed on this point. 

\section{Conclusion}

The present work addresses the fact, introduced in prior work \cite{Gonen2019LipstickOA}, that gendered terms remain clustered in the embedding space of debiased word embeddings. We propose a two-step pipeline solution to combat this issue. Our pipeline combines a post-processing step ---MDR \cite{Hasan2017WordRV}
and a debiasing method---Cluster Debias. It is able to outperform state-of-art debias methods on mitigating bias on the measures proposed by Gonen and Goldberg \cite{Gonen2019LipstickOA}. The success of our pipeline also validates our proposed reasons behind the observations made by Gonen and Goldberg. First, that there existed a mismatch between the direction that gendered terms distributed along and the direction that debiasing methods remove. And second, that the non-linear classifier (e.g. SVM) is able to separate gendered words from manifold structure in the high dimensional word embedding space. 

We also test our pipeline on downstream tasks. We find that our model outperforms existing approaches on the coreference resolution tasks in terms of mitigating gender bias. However, critically, the improvement seen is not nearly as stark as our improvement over prior methods on the upstream bias tasks we consider.  This indicates that word embeddings encode gender bias in \emph{still} other ways, not necessarily captured by the cluster-based measures from prior work.  As such, in order to avoid a ``whack-a-mole'' approach for mitigating bias, we encourage a focus on the development of more downstream tasks, relative to further upstream analysis. 
\small
\bibliographystyle{splncs04}
\bibliography{paper.bib}
%






\end{document}